\RequirePackage{fix-cm} 
\documentclass[smallextended]{svjour3}       % onecolumn (second format)
\smartqed  % flush right qed marks, e.g. at end of proof
\usepackage{graphicx,amsmath, geometry}
\usepackage{hyperref}
\usepackage{caption}
\usepackage{placeins}
\usepackage{amsfonts}
\usepackage{booktabs}
\usepackage{xcolor}
\usepackage{subcaption}
\usepackage{marginnote}
\usepackage[colorinlistoftodos,prependcaption]{todonotes}\reversemarginpar
\usepackage{algpseudocode}
\usepackage{algorithm}
\begin{document}

\title{Accurate Prediction and Estimation of 3D-Repetitive-Trajectories using Kalman Filter, Machine Learning and Curve-Fitting Method}
\subtitle{}
\author{Aakriti Agrawal, Aashay Bhise, Rohitkumar Arasanipalai, Lima Agnel Tony, Shuvrangshu Jana and Debasish Ghose}
\institute{ $^*$ Guidance Control and Decision Systems Laboratory \\
              Department of Aerospace Engineering\\
              Indian institute of Science\\
              Bangalore-12, India\\         %  \\
}

\date{Received: date / Accepted: date}

\maketitle

\begin{abstract}
Accurate estimation and prediction of trajectory is essential for the capture of any high speed target. % In this paper, an extended Kalman filter is used to estimate the current location of target from its visual information and then predict its future position by using the observation sequence. 
In this paper, an extended Kalman filter (EKF) is used to track the target in the first loop of the trajectory to collect data points and then a combination of machine learning with least-square curve-fitting is used to accurately estimate future positions for the subsequent loops. The EKF estimates the current location of target from its visual information and then predicts its future position by using the observation sequence. %Target motion model is developed considering the approximate known pattern of the target trajectory. In this work, 
We utilise noisy visual information of the target from the three dimensional trajectory to carry out the predictions. The proposed algorithm is developed in ROS-Gazebo environment and is implemented on hardware.

\end{abstract}

\section{Introduction}
\label{intro}

Target capture is a challenging problem in robotics and is relevant to several applications spanning from anti-drone systems\cite{Ref0} to fruit picking. The target under consideration would be stationary or moving. When the target has a speed advantage, its future location estimation becomes difficult. Therefore, the estimation of target's location will aid in precise capture with minimal control effort. The primary motivation of performing the position estimation and trajectory prediction of the target is to facilitate predictive guidance that optimises the control effort. The proposed framework is also robust to intermittent information supply due to the target moving at high speeds. Furthermore, the interception strategy can be modified if the target is known to follow a fixed trajectory repetitively.

Several interesting works have been reported in literature about high speed target interception. 
The improvement of tracking/interception performance using several methods, is reported in \cite{Ref1}, \cite{Ref2}, \cite{Ref3}. Predictive guidance and learning based guidance are also proposed to improve the interception performance, as seen in \cite{Ref4}, \cite{Ref5}, \cite{Ref6}, \cite{Ref7}. Methods pertaining to the field of soft computing applied in predictive guidance also provide promising results as seen in \cite{Ref8}, \cite{Ref9}, \cite{Ref10}. In existing literature, the target motion model is considered in general and formulation to include the known approximate repetitive motion of target using visual information is not reported. We are using several looping shapes as our target trajectory, one of which is the figure of 8 present in challenge 1 of MBZIRC 2020.

In this paper, we present the framework which is designed to estimate and predict the position of a moving target, which follows a repetitive path of some standard shape. While formulating the motion model for target position estimation, the following assumptions regarding the target motion are made. The motion of the target is assumed to be smooth i.e., the change in curvature of the target's trajectory remains bounded and smooth over time. This assumption is the basis of our formulation of the target motion model. The measurement sensor in this case is the vision module. The vision module uses image processing algorithms to compute the estimated position and velocity of the target in inertial plane.  

In the following section, a detailed mathematical formulation of the EKF method is provided. Following that, the least squares curve fitting method is described, both in 2D and 3D. It is then followed by the simulation results. Finally, conclusions and future work is described.

% Different sections: 
% a)  Description about estimation (visual) \\
% b)  Some description from vision about estimation \\
% c)  Prediction formulation EKF to track the object\\
% c1) Algorithm flow chart
% c2) Estimation using RNN 
% c3) The information from both is fused to predict the trajectory of target. 
% d)  Estimation of Curve using curve fitting-2D\\
% e) Curve fitting in 3D - flow chart if possible\\
% g) Interception approach based on curve (not included in both the curves, simulation performed as I remember)\\ 
% g1) Algorithm flow chart (complete algorithm)
% f) Simulation: \\
% 1)  Clear comparison between the actual position and predictive position\\
% 2) Curve fitted vs actual curve (2D/ 3D)\\
% 3) Complete simulation for approach points\\
% g) Description about testing setup \\
% h) Hardware testing results \\
% i) Conclusions

\section{Mathematical Formulation for EKF Method}
While formulating the motion model for target position estimation, the following assumptions regarding the target motion are made. The motion of the target is assumed to be smooth i.e., the change in curvature of the target's trajectory remains bounded and smooth over time. This assumption is the basis of our formulation of the target motion model.

It is assumed that target is maneuvering in a plane. The inertial position coordinates of the target $ (p_{n}, p_{e}) $ are considered as variables of state vector. They are measured using the vision module as the positional information of target obtained from camera frame but transformed into the inertial frame, denoted as $ (p_{e_{\text{image}}}, p_{n_{\text{image}}}) $. So, the state $x$ and measurement $y$ variable  are written in as per the notations in \eqref{eq:PE_state} and \eqref{eq:PE_measurement}, respectively.
% \todo[inline]{Add rephrased intro to this section}

\begin{equation}\label{eq:PE_state}
    x = \begin{bmatrix} 
    p_e & p_n \\
    \end{bmatrix} ^T
\end{equation} 

\begin{equation}\label{eq:PE_measurement}
    y =
    \begin{bmatrix} 
    p_{e_{\text{image}}} & p_{n_{\text{image}}} \\
    \end{bmatrix} ^T
\end{equation}

\begin{equation}\label{eq:PE_input}
    u = 
    \begin{bmatrix} 
    V_a &
    p_{e_0} &
    p_{n_0} \\
    \end{bmatrix} ^T
\end{equation}

An Extended Kalman Filter is used for the target's position estimation. The formulation is done with the consideration that the target's trajectory is lying in the inertial $X-Y$ plane; the $X-Y$ plane being in the E-N direction. The state vector contains the east and north positions. The input vector $u$  contains the estimated speed of the target as given by the vision module and co-ordinates of the center of the trajectory's instantaneous curvature. The measurement model \eqref{eq:PE_measurement} contains the position of the target in inertial $X-Y$ plane, as given by the vision module. The co-ordinates of the center of the instantaneous curvature are calculated by estimating the evolution matrix in the Least Square method. Evolution matrix is formulated by writing the future states of the target as a function of current state. The governing relations are as below.
%\todo[inline]{Expand on the evolution matrix}

The target motion is formulated as in equation \eqref{eq:PE_LS_motion_model} where $r$ is the radius of the instantaneous circle and $\delta$ is the change in $\theta$ between the timesteps. The future states are expressed as a function of previous states as shown in equation \eqref{eq:PE_LS_delta_state}, where $j$ is the index of the observations. The system of equations with the evolution matrix  (\( [ \cos\delta \ \sin\delta ]^T \)) is shown in \eqref{eq:PE_LS_matrix_eq}. A sequence of observation is gathered which fills the matrix equation. The Least Squares solution of the observation sequence provides the estimation of evolution matrix at every sample step, so that co-ordinates of the center of curvature, i.e $ (p_{e_0},p_{n_0}) $ in  \eqref{eq:PE_LS_center}, is available at every time step.    
% \todo[inline]{Figure of the instantaneous circle if possible}

\begin{equation}\label{eq:PE_LS_motion_model}
\begin{split}
    p_e(k+1) = p_e(k) - r\delta \sin\theta(k)  \\
    p_n(k+1) = p_n(k) + r\delta \cos\theta(k)
\end{split}
\end{equation}

\begin{equation}\label{eq:PE_LS_delta_state}
\begin{split}
    \Delta_{p_e}(k,j) = p_e(k-j) - p_e(k-j-1) = -r\delta \sin\theta(k-j-1)  \\
    \Delta_{p_n}(k,j) = p_n(k-j) - p_n(k-j-1) = r\delta \cos\theta(k-j-1)
\end{split}
\end{equation}

\begin{equation}\label{eq:PE_LS_matrix_eq}
\begin{bmatrix} 
    \vdots  \\
    \Delta_{p_e}(k,j) \\
    \Delta_{p_n}(k,j) \\
    \vdots 
    \end{bmatrix}
    =
    \begin{bmatrix} 
    \cos\delta \\
    \sin\delta
    \end{bmatrix}
    \begin{bmatrix} 
    \vdots & \vdots \\
    \Delta_{p_e}(k,j-1) & -\Delta_{p_n}(k,j-1) \\
    \Delta_{p_n}(k,j-1) & \Delta_{p_e}(k,j-1)  \\
    \vdots & \vdots
    \end{bmatrix}
\end{equation}

\begin{equation}\label{eq:PE_LS_center}
\begin{split}
    p_{e_0} = p_e(k) - \frac{\Delta_{p_n}(k+1)}{\delta} \\
    p_{n_0} = p_n(k) + \frac{\Delta_{p_e}(k+1)}{\delta}
\end{split}
\end{equation}

The EKF formulation and algorithm are very well known. We detail our formulation of prediction model and measurement model to convey important implementation details. Here the goal is to estimate the position in inertial frame (i.e., ENU frame) along the $ X $ and $ Y $ direction. The 2-dimensional vector $ x $, as shown in equation \eqref{eq:PE_state},  comprises of the positions in inertial frame. A 3-dimensional vector comprising of speed of target $V_a$ and the instantaneous center co-ordinates in inertial frame ( $p_{n_0}$ and $p_{e_0}$)is fixed as the input vector, as shown in equation \eqref{eq:PE_input}. The process can be described as a non-linear system with,
% \todo[inline]{QUE: Need to show EKF background? Looks unnecessary. JB will give sample}

\begin{equation}\label{eq:state_eq}
 \dot{x} = F(x,u) + \xi
\end{equation}
\begin{equation}\label{eq:PE_state_trans}
    F(x,u) = 
    \begin{bmatrix} 
     - V_a( p_n - p_{n_0} )/\sqrt{( p_n - p_{n_0} ) ^2 + ( p_e - p_{e_0} ) ^2} \\
    V_a( p_e - p_{e_0} )/\sqrt{( p_n - p_{n_0} ) ^2 + ( p_e - p_{e_0} ) ^2} 
    \end{bmatrix} 
\end{equation}

\begin{equation}\label{eq:meas_eq}
y = H(x,u) + \eta 
\end{equation}

\begin{equation}\label{eq:PE_measure_model}
H(x,u) =
\begin{bmatrix} 
p_e\\
p_n
\end{bmatrix} 
\end{equation}

where, in equation \eqref{eq:state_eq}, $ F(x,u) $ (refer to equation \eqref{eq:PE_state_trans}) is the non-linear state transition function and $  \xi \sim \mathcal{N}(0,Q) $ is the process noise where the covariance $ Q $ is generally unknown and can be tuned. The process noise is assumed to be normally distributed. The measurement space contains 2 measurements of $ p_e $ and $ p_n $ in inertial frame, as shown in equation \eqref{eq:PE_measurement}. The measurement model is of the form \eqref{eq:meas_eq} where, $ H(x,u) $ (refer to equation \eqref{eq:PE_measure_model}) is the non-linear measurement model which maps state and input into measurement space and $  \eta \sim \mathcal{N}(0,R) $ is the measurement noise where the covariance $ R $ can be estimated by calibrating the sensors. The measurement noise is also assumed to be normally distributed. 
% \todo[inline]{MINOR: add a line about the process noise}

Prediction step is the first stage of the EKF algorithm where we propagate the previous state and input values to the non-linear process equation \eqref{eq:EKF_prop} in discrete time estimate to arrive at the state estimate. The error covariance matrix is projected by the Jacobian of state transition function \eqref{eq:EKF_state_Jac} and perturbed by the process noise covariance, as shown in equation \eqref{eq:EKF_pred_step_cov}.

\begin{equation}\label{eq:EKF_prop}
\dot{\hat{x}} = F(\hat{x},u) 
\end{equation}

\begin{equation}\label{eq:EKF_state_Jac}
    A = \frac{\partial F(\hat{x},u)}{\partial x} = \frac{V_a}{(( p_n - p_{n_0} ) ^2 + ( p_e - p_{e_0} ) ^2)^{3/2} }
    \begin{bmatrix} 
    ( p_n - p_{n_0} )( p_e - p_{e_0} ) & -( p_e - p_{e_0} )^2 \\
    ( p_n - p_{n_0} )^2 & -( p_e - p_{e_0} )( p_n - p_{n_0} ) 
    \end{bmatrix} 
\end{equation}

\begin{equation}\label{eq:EKF_pred_step_cov}
\dot{P} = AP + PA^T + Q 
\end{equation}

The correction step is then carried out after the measurement update where we calculate the Jacobian of the measurement model \eqref{eq:EKF_meas_Jac} and later the Kalman Gain $L$ in \eqref{eq:EKF_kalman_gain}. Then eventually the state estimate \eqref{eq:EKF_state_corr} and the error covariance matrix \eqref{eq:EKF_corr_step_cov} is updated. This is the underlying mathematical formulation of the position estimation framework.

\begin{equation}\label{eq:EKF_meas_Jac}
   C =  \frac{\partial H(\hat{x},u)}{\partial x} =
    \begin{bmatrix} 
    1 & 0\\
    0 & 1
    \end{bmatrix} 
\end{equation}

\begin{equation}\label{eq:EKF_kalman_gain}
L = PC^T(R + CPC^T)^-1 
\end{equation}

\begin{equation}\label{eq:EKF_corr_step_cov}
P = (I - LC)P 
\end{equation}

\begin{equation}\label{eq:EKF_state_corr}
\dot{\hat{x}} = L(y - H(\hat{x},u)) 
\end{equation}

The later part of the task includes predicting the future state of the target based on a sequence of filtered states of target. The position estimation framework provides a filtered position of the target, which is then used for predicting the trajectory of the target. The workflow of trajectory prediction is divided into two, namely, observation phase and prediction phase. During the observation phase, a predefined sequence of observations are gathered \eqref{eq:TP_LS_delta_state}. These observations are the estimated position of the target in inertial frame. Evolution matrix is calculated by obtaining the Least squares solution over the gathered sequence \eqref{eq:TP_LS_matrix_eq}. The evolution matrix is then propagated in the motion model to predict the trajectory in near future. The formulation of the motion model is given in equation \eqref{eq:TP_LS_motion_model}. 
% \todo[inline]{Add a figure for traj pred framework}

\begin{equation}\label{eq:TP_LS_motion_model}
\begin{split}
    p_e(k+1) = p_e(k) - (t_{k+1} - t_k)V_a \sin\theta(k)  \\
    p_n(k+1) = p_n(k) + (t_{k+1} - t_k)V_a \cos\theta(k)
\end{split}
\end{equation}

\begin{equation}\label{eq:TP_LS_delta_state}
\begin{split}
    \Delta_{p_e}(k,j) = p_e(k-j) - p_e(k-j-1) = -(t_{k+1} - t_k)V_a \sin\theta(k-j-1)  \\
    \Delta_{p_n}(k,j) = p_n(k-j) - p_n(k-j-1) = (t_{k+1} - t_k)V_a \cos\theta(k-j-1)
\end{split}
\end{equation}

\begin{equation}\label{eq:TP_LS_matrix_eq}
\begin{bmatrix} 
    \vdots \\
    \Delta_{p_e}(k,j) \\
    \Delta_{p_n}(k,j) \\
    \vdots
    \end{bmatrix}
    =
    \begin{bmatrix} 
    \cos\delta \\
    \sin\delta
    \end{bmatrix}
    \begin{bmatrix} 
    \vdots & \vdots \\
    \Delta_{p_e}(k,j-1) & -\Delta_{p_n}(k,j-1)\\
    \Delta_{p_n}(k,j-1) & \Delta_{p_e}(k,j-1) \\
    \vdots & \vdots
    \end{bmatrix}
\end{equation}

for j = 1,2,$ \cdots $,m

\begin{equation}\label{eq:pred_phase}
\begin{split}
\Delta_{p_e}(k,j) = -(t_{k+1} - t_k)V_a (\sin\theta(k+j-2) + \delta)  \\
\Delta_{p_n}(k,j) = (t_{k+1} - t_k)V_a (\cos\theta(k+j-2) + \delta)
\end{split}
\end{equation}

\begin{equation}\label{eq:pred_phase_prop}
\begin{split}
\hat{p_{e}}(k+j|k)=\hat{p_{e}}(k)+\Delta_{p_e}(k,j) \\
\hat{p_{n}}(k+j|k)=\hat{p_{n}}(k)+\Delta_{p_n}(k,j)
\end{split}
\end{equation}

The formulation of the motion model is given in equation \eqref{eq:TP_LS_motion_model}. In the prediction phase the targets future positions within $ m $ steps is estimated by the latest evolution matrix obtained from \eqref{eq:TP_LS_matrix_eq}. The evolution matrix is then propagated \eqref{eq:pred_phase_prop} in the motion model to predict the trajectory in near future. This concludes the joint framework of the EKF based target position estimation and Least Squares based future trajectory prediction.

% \todo[inline]{SUMMARY: Brief overview + vision data transf to inertial frame.
% 	Target state estimation from vision followed by future trajectory prediction.
% 	Flow chart of the process.}

% FO8 PARAM. EST. START
\color{black}
\section{Mathematical Formulation for Curve Fitting Method}

The EKF method does not use any prior set of data about the target trajectory. Therefore, it's usage is ideal in the first loop. After the first loop is done, we use the data collected to give a highly accurate estimate about the target trajectory.  

Assumptions: The target drone is moving autonomously in a looping trajectory. Therefore, we can conclude that it will be following a trajectory whose curve-equation is mathematically derivable. Also, it will be a closed loop curve since the trajectory is repetitive. We have taken into consideration all high-order closed curves to the best of our knowledge and curve fit the data to the appropriate curve-equation without prior knowledge about the shape of figure. 

The closed curves taken into consideration are listed in Table \ref{tab:curves}. This method will also work with any closed mathematically-derivable curve. Since, circle is a special case of ellipse we will include both in a single category.

\begin{table}
  \centering
    \begin{tabular}{|l|l|}
        \hline
        \text{Curves} & \text{Curve Equation}\\
        \hline
        \text{Circle/Ellipse} & $\frac{x^2}{a^2} + \frac{y^2}{b^2} = 1$ \\
        \text{Astroid} & $x^{\frac{2}{3}} + y^{\frac{2}{3}} = a^{\frac{2}{3}}$\\
        \text{Deltoid} & $(x^2 + y^2)^2 + 18a^2(x^2 + y^2) - 27a^4 = 8a(x^3 - 3xy^2)$ \\
        \text{Limacon} &  $(x^2 + y^2 - ax)^2 = b^2(x^2 + y^2)$\\
        \text{Nephroid} & $(x^2 + y^2 - 4a^2)^3 = 108a^4y^2$ \\
        \text{Quadrifolium} & $(x^2 + y^2)^3 = (x^2 - y^2)^2$ \\
        \text{Squircle} & $(x-a)^4 + (x-b)^4 = r^4$ \\
        \text{Lemniscate of Bernoulli} & $(x^2 + y^2)^2 = 2a^2(x^2 - y^2)$ \\
        \text{Lemniscate of Gerono} &  $x^4 = a^2(x^2 - y^2) $\\
        \hline
    \end{tabular}
    \caption{Equations of the all high order curves taken into consideration.}
        \label{tab:curves}
\end{table}
The curves mentioned above have been well studied and their characteristics are well known. They are usually the zero set of some multivariate polynomials. We can write their equations as 
\begin{equation}\label{eq:gen_func}
    f(x,y) = 0 
\end{equation}
For example, the leminscate of bernoulli is
\begin{equation}\label{eq:func_gerono}
    (x^2 + y^2)^2 - 2a^{2}(x^2 - y^2) = 0 
\end{equation}
Leminscate of bernoulli has a single parameter $a$, which needs to be estimated. The equation of ellipse, on the other hand, has 2 parameters $a$ and $b$ that needs to be estimated. Therefore, we can write a general function for the curves as 
\begin{equation}\label{eq:gen_func}
    f(x, y, a, b) = 0 
\end{equation}
where $b$ may or may not be used based on the category of shape the points are being fitted to.

Univariate polynomials of the form
\begin{equation}
    f(x) = a_{0} + a_{1}x + a_{2}x^2 + ... + a_{k}x^k
\end{equation}
can be solved using matrices if there are enough points to solve for the $k$ unknown coefficients. On the other hand, multivariate equations require different methods to solve for their coefficients. One method for curve fitting uses an iterative least-squares approach along with specifying related constraints.

\subsection{Classification of curves}\label{sec:class_curve}
The above mentioned categories of curves all have different equations. There is requirement to classify the curve into one of the above categories before curve-fitting. We train a neural network to classify the curves into the various categories based on the $(x,y)$ points collected from the target drone.

\begin{figure}[h]
  \centering
  \includegraphics[width=0.5\linewidth]{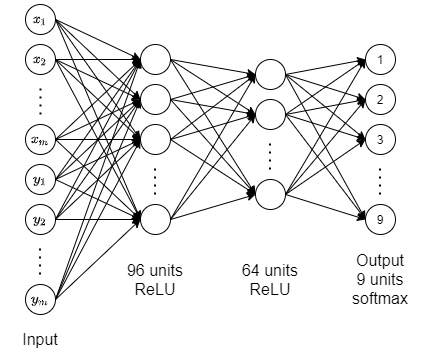}
  \caption{Network Architecture}
\label{fig:network_arch}
\end{figure}

The architecture of the network used is shown in Figure \ref{fig:network_arch}. The input $I$ is a vector of $m$ points arranged as $[x_0,x_1,\ldots,x_m,y_0,y_1,\ldots,y_m]$. The output $O$ is a vector of length 9 denoting the probabilities of the given set of points belonging to the various categories. Therefore, the network can be represented as a function $f$ trained to map
\begin{equation}
    f:[x_0,x_1,\ldots,x_m,y_0,y_1,\ldots,y_m] \mapsto O
\end{equation}
The training parameters are listed in Table \ref{tab:optimizer_specs}. 
\begin{table}
  \centering
    \begin{tabular}{|l|l|}
        \hline
        \text{Parameters} & \text{Value}\\
        \hline
        \textbf{Optimizer} &  Adam\\
        \textbf{Learning Rate} & $10^{-4}$\\
        \textbf{No. of Training Epochs} & 9 \\
        \textbf{Final Training Accuracy} & 98\% \\
        \hline
    \end{tabular}
    \caption{Training parameters for the classification network}
    \label{tab:optimizer_specs}
\end{table}

This network can classify 2D curves into the above mentioned categories. In case of 3D, we can use this same network to classify the curve once it has been rotated into a 2D plane (like the $X-Y$ plane).

\subsection{Least-Squares Curve Fitting in 2D}
\label{method:fo8-2D}
Considering any of the above mentioned curves in two dimensions, the base equation has to be modified to account for both offset and orientation in 2D. Therefore, let the orientation be some $\theta$, and the offset be $(x_0,y_0)$.
On applying a counter-clockwise $\theta$ rotation to a set of points, the rotation is defined by this matrix equation:
\begin{equation}\label{eq:rot}
    \begin{bmatrix} x \\ y \end{bmatrix} = 
    \begin{bmatrix} \cos\theta & -\sin\theta \\
    \sin\theta & \cos \theta \end{bmatrix}
    \begin{bmatrix} x' \\ y' \end{bmatrix}
\end{equation}
Substituting $(x,y)$ from Equation \ref{eq:rot} into Equation \ref{eq:gen_func}, we get the following function:

% \begin{multline}\label{eq:complex}
%     2a^2((y\cos\theta + x\sin\theta)^2 - (x\cos\theta - y\sin\theta)^2) + \\ ((y\cos\theta + x\sin\theta)^2 + (x\cos\theta - y\sin\theta)^2)^2 = 0
% \end{multline}
\begin{equation}\label{eq:complex}
    f(y'\cos\theta + x'\sin\theta,x'\cos\theta - y'\sin\theta, a, b) = 0
\end{equation}
Letting 
\begin{equation}
    g(x',y',\theta, a, b) = f(y'\cos\theta + x'\sin\theta,x'\cos\theta - y'\sin\theta, a, b)
\end{equation}
and rewriting it by replacing $x'$ and $y'$ by $x$ and $y$ repectively, we have 
\begin{equation}\label{eq:complex_g_rot}
    g(x,y,\theta, a, b) = 0
\end{equation}

To account for offset from origin, we can replace all $x$ and $y$ with $x'$ and $y'$, respectively, where,
\begin{align}
    x' &= x - x_{0} \\
    y' &= y - y_{0}
\end{align}
and $(x_{0},y_{0})$ is the offset of the centre of the figure from the origin. Therefore, we have
\begin{equation}\label{eq:complex_g_final}
    g(x,y,\theta, a, b, x_0, y_0) = 0
\end{equation}
as the final equation of the figure we are trying to fit.
Applying the least-squares method on above equation for curve-fitting of $m$ empirical points $(x_i, y_i)$.   
\begin{equation}
    E^2 = \sum_{i=0}^{m} (g(x_i, y_i, \theta, a, b, x_0, y_0)-0)^2
\end{equation}

Our aim is to find $x_0$, $y_0$, $a$, $b$ and $\theta$ such that $E^2$ is minimised. This can only be done by,
\begin{equation}\label{eq:derivative}
    \frac{dE^2}{d\beta} = 0, \text{where } \beta \in \{x_0, y_0, a, b \theta\}
\end{equation}
If $g$ had been a linear equation, simple matrix multiplication would have yielded the optimum parameters. But since equation \ref{eq:complex_g_final} is a complex $n^{\text{th}}$ (where $n$ is 2, 4 or 6) order nonlinear equation with trigonometric variables, we need to use iterative methods in order to estimate the parameters, $a$, $b$, $\theta$, $x_0$, and $y_0$. Therefore, this work uses Levenberg–Marquardt \cite{LM_L,LM_M} least-squares algorithm to solve the non-linear equation \ref{eq:derivative}.

\subsection{Least-Squares Curve Fitting of any shape in 3D}\label{3d_transf}

If the orientation of any shape is in 3D, the above algorithm will need some modifications. We first compute the equation of the plane in which the shape lies and then transform the set of points to a form where the method in Section \ref{method:fo8-2D} can be applied.

In order to find the normal to the plane of the shape, we carry out singular value decomposition (SVD) of the given points. Let the set of points ($x$, $y$, $z$) be represented as matrix $A$ $\in$ 	$\mathbb{R}^{n\times3}$. From each point subtract the centroid and calculate SVD of $A$.

\begin{equation}
    A = U \Sigma V.    
\end{equation}

where, columns of $U = (u_{1}, u_{2} ..... u_{n})$ (left singular vectors), span the space of columns of $A$, columns of $V = (v_{1}, v_{2}, v_{3})$ (right singular vectors) span the space of rows of $A$ and $\Sigma = diag(\sigma_{1}, \sigma_{2}, \sigma_{3})$ are the singular values linked to each left/right singular vector. Now, since the points are supposed to be in 2D space, $\sigma_{3} = 0$ and $v_{3} = (n_1,n_2,n_3)$ gives the normal vector to the plane. Therefore, the equation of the plane is,

\begin{equation}
    n_{1}x+n_{2}y+n_{3}z = C, \text{where C is a constant.} 
    \label{eq:plane}
\end{equation}

The next step is to transform the points to $X-Y$ plane. For that we first find the intersection of the above plane with x-y plane by substituting $z=0$ in equation \ref{eq:plane}. We get the equation of line as,

\begin{equation}
    n_{1}x+n_{2}y = C, \text{where C is a constant.} 
\end{equation}

Then we rotate the points about $z$-axis such that the above line is parallel to $x$-axis. Angle of rotation $\alpha = 0$, $\beta = 0$ and $\gamma = \arctan(\frac{n_1}{-n_2})$ needs to be substituted in matrix R, given in equation \ref{eq:Rotationmatrix_3D}. New points will be $A_{z} = A R$.  

\begin{equation}
\label{eq:Rotationmatrix_3D}
    R = 
    \begin{bmatrix}
        \cos\beta\sin\gamma \text{ }& \sin\alpha\sin\beta\cos\gamma - \cos\alpha\sin\gamma \text{ }& \cos\alpha\sin\beta\cos\gamma + \sin\alpha\sin\gamma \\
        \cos\beta\sin\gamma & \sin\alpha\sin\beta\sin\gamma + \cos\alpha\cos\gamma & \cos\alpha\sin\beta\sin\gamma - \sin\alpha\cos\gamma \\
        -\sin\beta & \sin\alpha\cos\beta & \cos\alpha\cos\beta
        
    \end{bmatrix}
\end{equation}

\begin{algorithm}
\caption{Least Means Squares Algorithm}\label{lms_algo}
\begin{algorithmic}[1]
    \State Initialize parameters for shape detection network
    \State Store datapoints in variable $Shape$
    \If {$Shape$ is in 3D}
    \State Transform the shape to the $X-Y$ plane using method in Section \ref{3d_transf}
    \State $Shape$ $\leftarrow$ $Shape_{transformed}$
    \EndIf
    \State Get shape prediction, $shape_{pred}$, of $Shape$ from shape detection network 
    \State Apply curve-fitting algorithms on $Shape$ using the equation of $shape_{pred}$ 
    \State Generate target drone trajectory using the estimated shape parameters
\end{algorithmic}
\end{algorithm}

% \subsection{Results for EKF formulation}\label{sec:results}

We then rotate the points about $X$-axis by the angle $\cos^{-1}(|n_{3}|/ \left\lVert n_{1},n_{2},n_{3} \right\rVert)$ to make the points lie in the $X-Y$ plane. Then, we substitute angles $\alpha = \arccos(\frac{|n_{3}|}{\left\lVert n_{1},n_{2},n_{3} \right\rVert})$, $\beta = 0$ and $\gamma = 0$ in the rotation matrix given in \ref{eq:Rotationmatrix_3D}. Finally, the set of points in the $X-Y$ plane will be $A_{final} = A_{z}R$. We can then use the neural network described in Section \ref{sec:class_curve} to classify the curve into one of the 9 categories. Then, we can compute the parameters ($x_{0}$, $y_{0}$, $a$, $b$, $\theta$) of the classified curve using method given in Section \ref{method:fo8-2D}. The combined algorithm for the shape detection and paramter estimation is shown in Algorithm \ref{lms_algo}.

% \subsubsection{Relating time with figure-of-8}

% \section{Combining EKF and Curve-Fitting}
% \todo[inline]{Should I add the EKF+LS flow chart here}

\section{Results}
\begin{figure}[h!]
    \centering
    \includegraphics[scale=0.2]{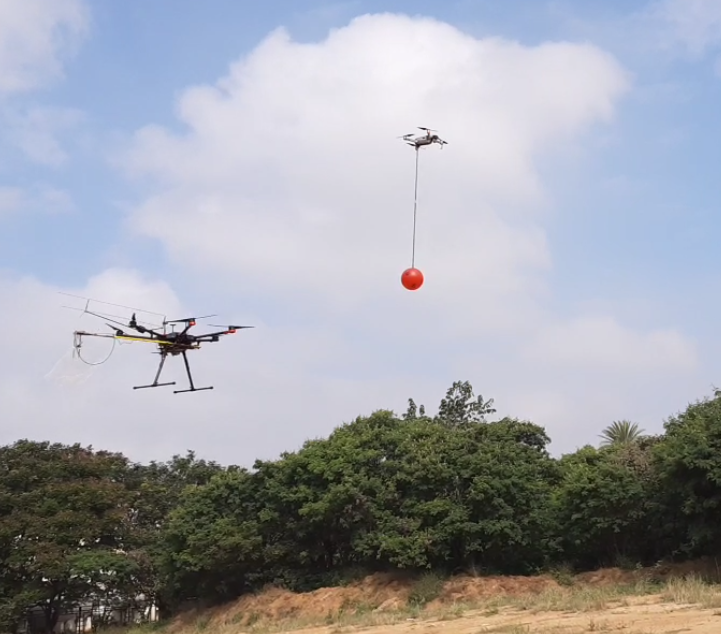}
    \caption{Own UAV and target UAV. Using visual information, own UAV estimates and predicts the location of target}
    \label{fig:UAVs}
\end{figure}
A ROS-based pipeline in written in C++ which performs target state filtering and future state estimation. The ROS-based packages are first tested in a gazebo environment where the information about the state of the target is obtained from a simulated drone 
% Fig. \ref{fig:gaz_env}.
Since the information about the state of the target is highly accurate and always available, the measurement co-variance is amplified and Gaussian noise is added to observed data, solely to create a realistic scenario. Position estimation of target is done over this noisy data and later, the future state of the target is predicted. A separate process sets the motion of the target in one of the various selected shapes. The states estimated and future states predicted are visualized in RViz. 
% As shown in Fig. \ref{fig:raw_filt_gaz}, 
The estimated data is visualized against the raw data of the various shapes. Here, since the estimated position is visualized in $X-Y$ plane, we restrict our prediction and estimation in two dimensions. 

\subsection{Least-Squares Curve Fitting Results}
\label{Results}

\begin{figure}[!b]
    \centering
    \begin{subfigure}{0.33\textwidth}
        \centering
        \includegraphics[width=\linewidth,trim={0cm 0cm 0cm 0cm},clip]{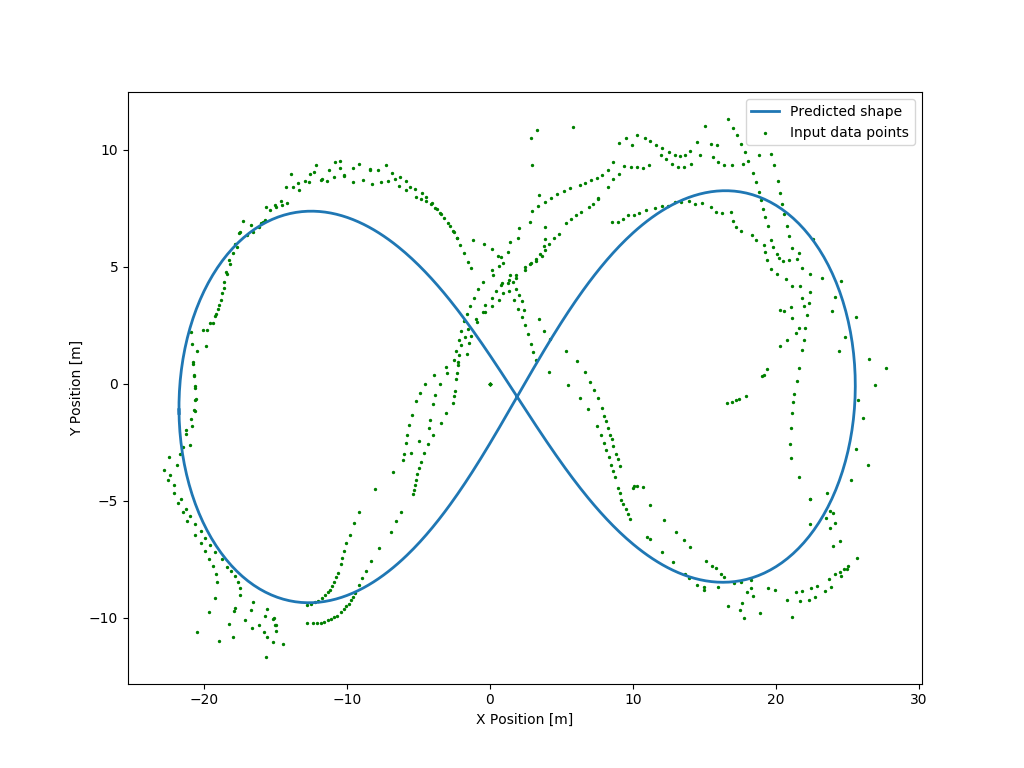}
        \caption{Leminscate of Bernoulli}
        \label{fig:ber}
    \end{subfigure}%
    ~ 
    \begin{subfigure}{0.33\textwidth}
        \centering
        \includegraphics[width=\linewidth,trim={0cm 0cm 0cm 0cm},clip]{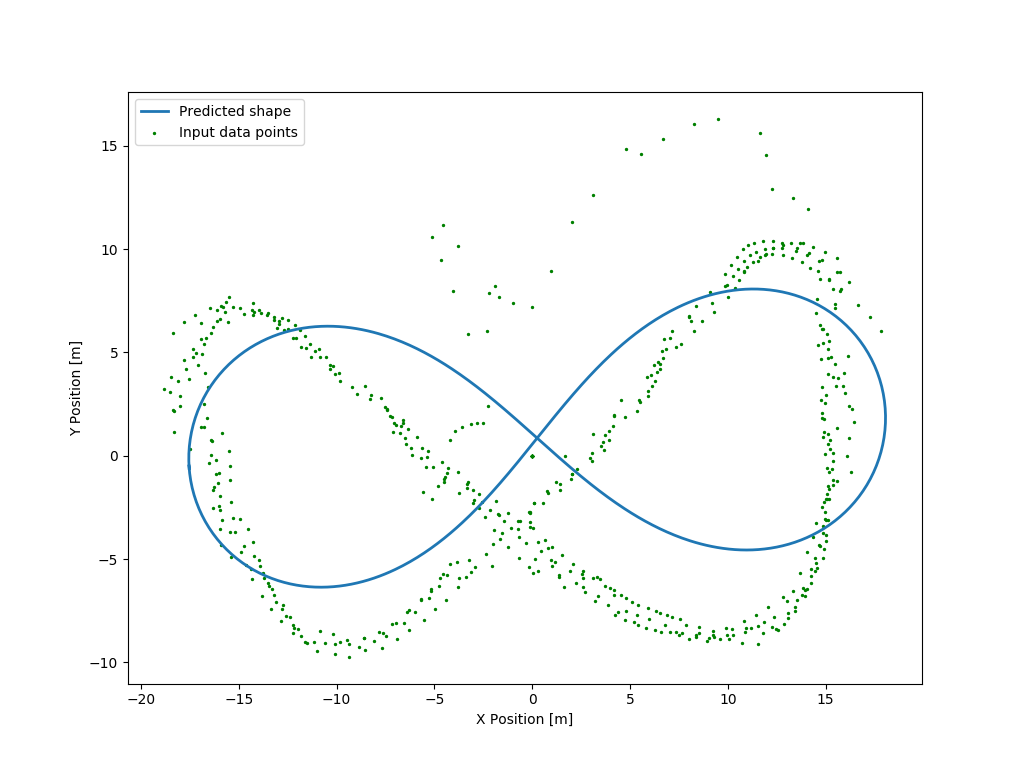}
        \caption{Leminscate of Genoro}
        \label{fig:gen}
    \end{subfigure}%
    ~
    \begin{subfigure}{0.33\textwidth}
        \centering
        \includegraphics[width=\linewidth,trim={0cm 0cm 0cm 0cm},clip]{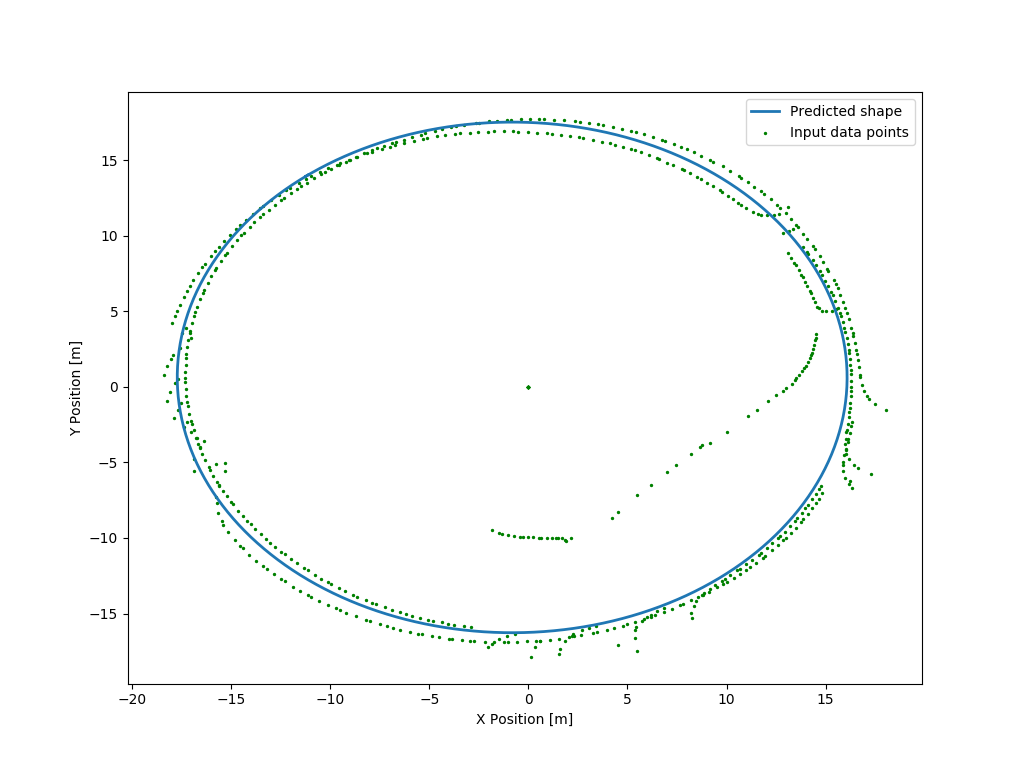}
        \caption{Circle/Ellipse (Second Degree Curves)}
        \label{fig:circle}
    \end{subfigure}
\end{figure}
\begin{figure}[!b]\ContinuedFloat
    \begin{subfigure}{0.33\textwidth}
        \centering
        \includegraphics[width=\linewidth,trim={0cm 0cm 0cm 0cm},clip]{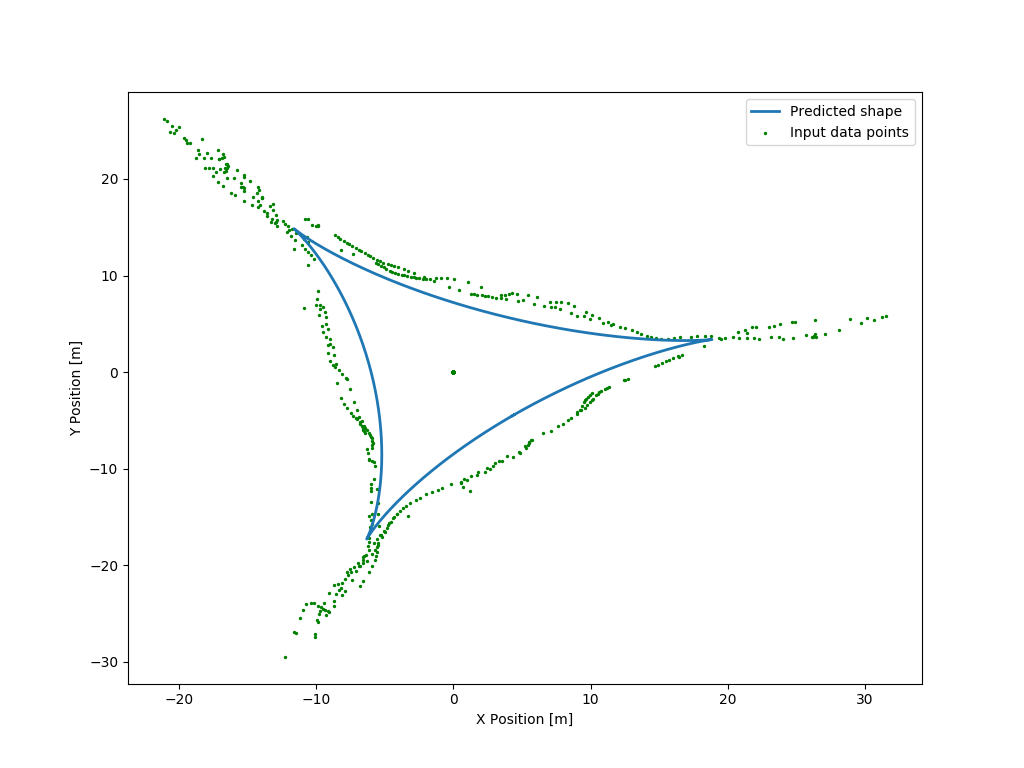}
        \caption{Deltoid}
        \label{fig:deltoid}
    \end{subfigure}%
    ~
    \begin{subfigure}{0.33\textwidth}
        \centering
        \includegraphics[width=\linewidth,trim={0cm 0cm 0cm 0cm},clip]{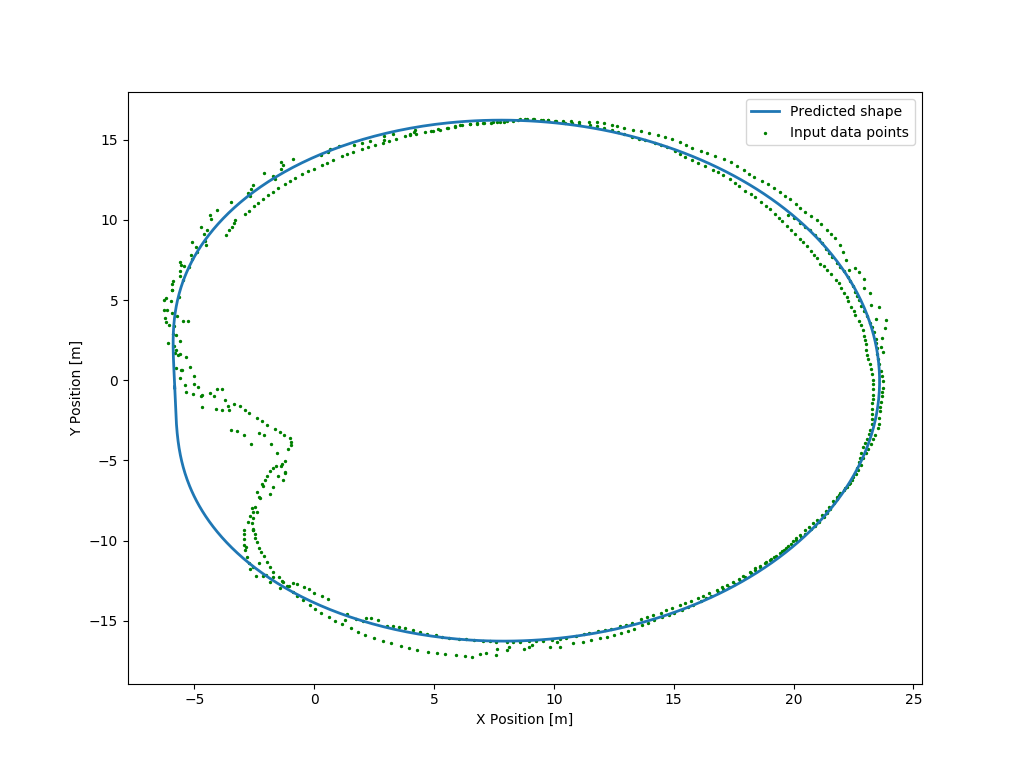}
        \caption{Limacon}
        \label{fig:limacon}
    \end{subfigure}%
    ~ 
    \begin{subfigure}{0.33\textwidth}
        \centering
        \includegraphics[width=\linewidth,trim={0cm 0cm 0cm 0cm},clip]{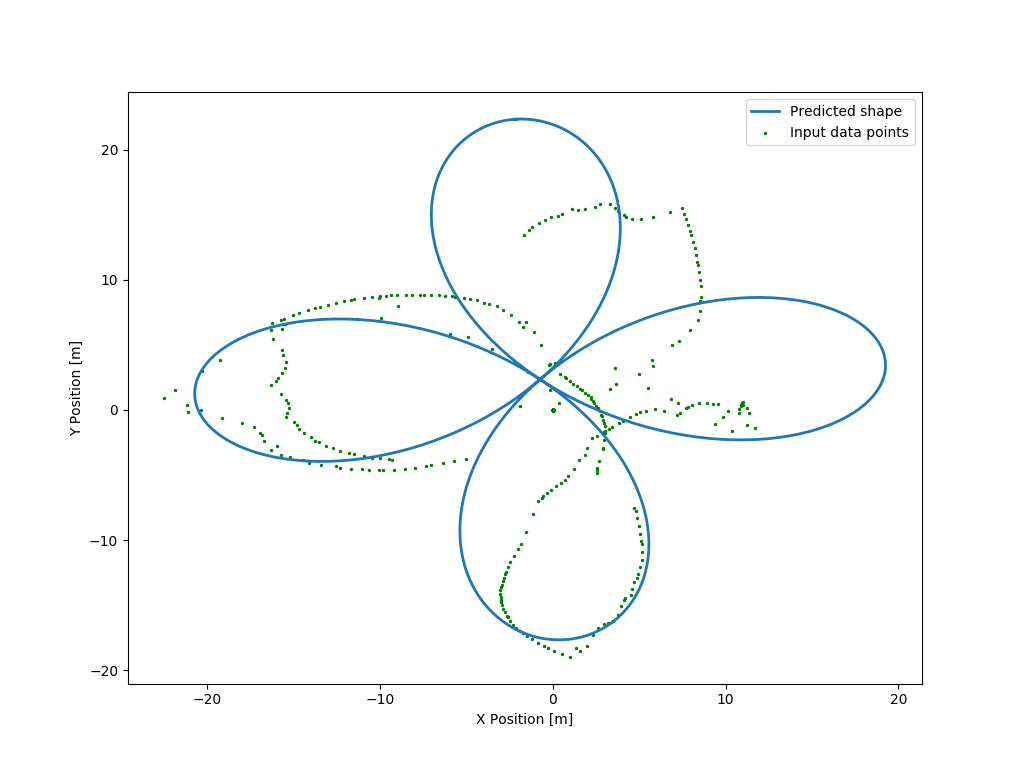}
        \caption{Quadrifolium}
        \label{fig:quadrifolium}
    \end{subfigure}
\end{figure}
\begin{figure}\ContinuedFloat
    \begin{subfigure}{0.33\textwidth}
        \centering
        \includegraphics[width=\linewidth,trim={0cm 0cm 0cm 0cm},clip]{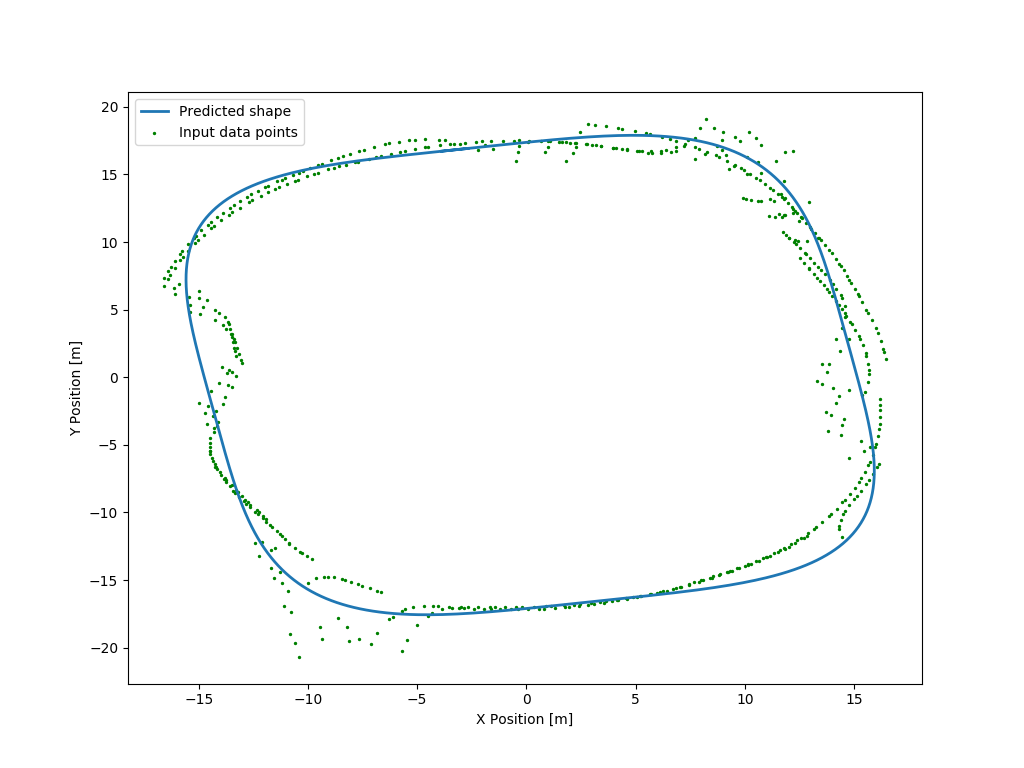}
        \caption{Squircle}
        \label{fig:squircle}
    \end{subfigure}%
    ~ 
    \begin{subfigure}{0.33\textwidth}
        \centering
        \includegraphics[width=\linewidth,trim={0cm 0cm 0cm 0cm},clip]{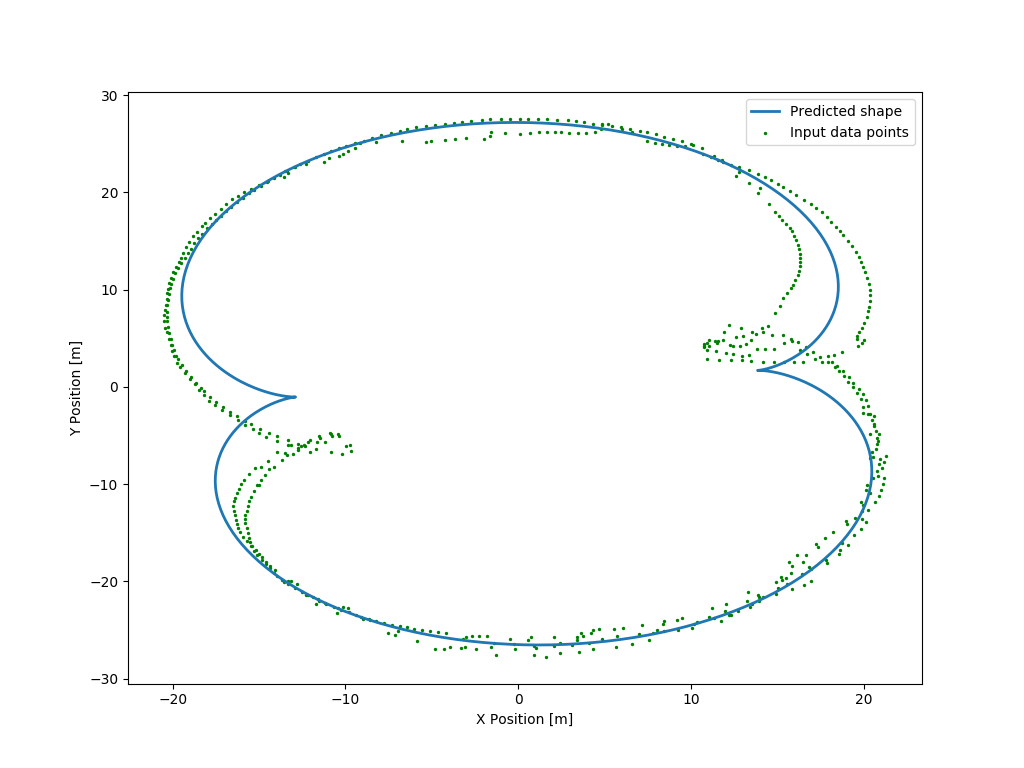}
        \caption{Nephroid}
        \label{fig:nephroid}
    \end{subfigure}%
    ~
    \begin{subfigure}{0.33\textwidth}
        \centering
        \includegraphics[width=\linewidth,trim={0cm 0cm 0cm 0cm},clip]{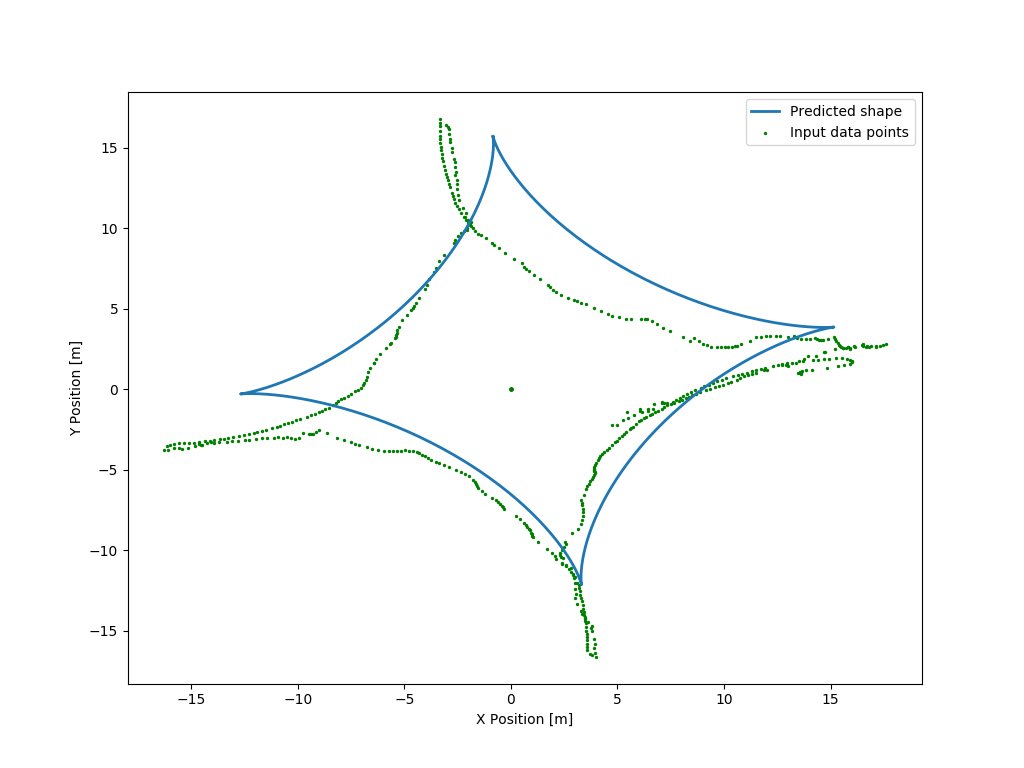}
        \caption{Astroid}%
        \label{fig:astroid}
    \end{subfigure}
    \caption{Figure showing the data points and predicted shape for the various shape categories}\label{fig:all_shapes}
\end{figure}

The results of the simulated data are shown in the following section. After one loop of data, the neural network predicts the category of the shape, after which the curve fitting algorithm estimates the shape parameters using the appropriate shape equation. As can be seen in Figure \ref{fig:all_shapes}, the curve fitting algorithm is able to accurately predict parameters for all of the shapes. 

\section{Conclusions}
In this paper, estimation of target location and future state prediction is performed using the  visual information. The proposed method is validated for target motion in a looping trajectory. Future work involves the prediction of target location following complex trajectory.

\begin{acknowledgements}
The authors would like to acknowledge support from Robert Bosch Center for Cyber Physical Systems, IISc, India.
\end{acknowledgements}

\end{document}